\documentclass[conference]{IEEEtran}
\IEEEoverridecommandlockouts

\usepackage{cite}
\usepackage{amsmath,amssymb,amsfonts}
\usepackage{algorithmic}
\usepackage{graphicx}
\usepackage{textcomp}
\usepackage{xcolor}
\usepackage{hyperref}
\usepackage{xspace}
\usepackage{amsmath}
\usepackage{placeins}
\usepackage[light]{antpolt}
\usepackage[T1]{fontenc}
\usepackage[margin=0.55in]{geometry}
\usepackage{pgfplots}
\pgfplotsset{width=10cm,compat=1.9}
\usetikzlibrary{external}
\usepgfplotslibrary{external}

\hypersetup{
    colorlinks=true,
    linkcolor=blue,
    filecolor=magenta,      
    urlcolor=cyan,
    pdftitle={},
    pdfpagemode=FullScreen,
    }

\begin{document}

\title{Classifying Textual Data with pretrained Vision Models through Transfer Learning and Data \\
Transformations
}

\author{\IEEEauthorblockN{Charaf Eddine Benarab}
\IEEEauthorblockA{\textit{University of Electronics Science and Technology of China} \\
\textit{School of Computer Science and Engineering}\\
Chengdu, China \\
charafeddineben@std.uestc.edu.cn}
}

\maketitle

\begin{abstract}
Knowledge is acquired by humans through experience, and no boundary is set between the kinds of knowledge or skill levels we can achieve on different tasks at the same time. When it comes to Neural Networks, that is not the case. The breakthroughs in the field are extremely task and domain-specific. Vision and language are dealt with in separate manners, using separate methods and different datasets. Current text classification methods, mostly rely on obtaining contextual embeddings for input text samples, then training a classifier on the embedded dataset. Transfer learning in Language-related tasks in general, is heavily used in obtaining the contextual text embeddings for the input samples. 
In this work, we propose to use the knowledge acquired by benchmark Vision Models which are trained on ImageNet to help a much smaller architecture learn to classify text. A data transformation technique is used to create a new image dataset, where each image represents a sentence embedding from the last six layers of BERT, projected on a 2D plane using a t-SNE based method. 
We trained five models containing early layers sliced from vision models which are  pretrained on ImageNet, on the created image dataset for the IMDB dataset embedded with the last six layers of BERT.
Despite the challenges posed by the very different datasets, experimental results achieved by this approach which links large pretrained models on both language and vision, are very promising, without employing compute resources. Specifically, Sentiment Analysis is achieved by five different models on the same image dataset obtained after BERT embeddings are transformed into gray scale images. 
\end{abstract}


\section{Introduction}
Attention-based architectures and specifically the Transformer~\cite{attention} sparked a revolution in the world of Natural Language Processing and Deep Learning in General. BERT-representations \cite{Bert} opened a lot of new goals and challenges for the Machine Learning Community, because of their semantically rich embeddings- and the kind of knowledge they encode given textual data. Being pretrained to Masked Language and Next Sentence prediction and on a very large Wikipedia Corpus, makes it a powerful benchmark for Natural Language Processing and a current standard for word and sentence representations. Computer Vision made very remarkable achievements~\cite{visionsurvey}. The groundbreaking work introduced in AlexNet~\cite{alexnet} and the parallel structure of Convolutional Neural Networks enabled the use of GPUs, making it a standard for training Neural Networks for Visual Recognition. Different architectures have emerged~\cite{alexnet}~\cite{shufflenetv2}~\cite{vgg16}~\cite{resnet}~\cite{resnext} since then using CNN's as a building block and achieving very high accuracies on different datasets. Transfer Learning~\cite{transferlearning}~\cite{transferlearningreminder} opened the doors for a very wide range of applications, allowing for the exchange of previously acquired knowledge from a large dataset on a certain task to another task with a much smaller dataset. This philosophy is often an essential practice in both academia and industry as it helps avoid a very long and computationally demanding procedure. In visual understanding tasks, transfer learning in the last decade heavily relied on ImageNet~\cite{imagenet} as a base or source dataset, ImageNet~\cite{imagenet} is a large dataset containing over 14 million images with annotations from one-thousand class labels and is often selected for pre-training Vision Models. The work described in this paper aims to use knowledge acquired by vision models to train text-classifiers for Sentiment Analysis, using a t-SNE based transformation method brought about in \cite{deepinsight} to transform IMDB Text Embeddings from BERT into gray scale images. The main objective of this paper is to bring language and vision a step closer, and to harness the power of transfer learning from large image datasets in Natural Language Processing and vice-versa. We hope this will encourage further work on the topic using appropriate resources and datasets since none were available during the conception of this paper. We extend the relevant body of work on the integration of language and vision covered in [18] with a complete Transfer Learning-based fusion of these modalities. The main components of our study include:

\begin{itemize}
    \item Using the BERT~\cite{Bert} embeddings for the IMDB-Dataset~\cite{imdb} to create
    an IMDB-Image Dataset using the method described in \cite{deepinsight} with a t-SNE~\cite{tsne} backbone. 
    \item Analyzing Domain Shifts between the Source (ImageNet~\cite{imagenet}) and Target (IMDB-Image) datasets, and avoiding such problems with feature normalization.
    \item Using early layers from benchmark Vision Models~\cite{alexnet}~\cite{shufflenetv2}~\cite{vgg16}~\cite{resnet}~\cite{resnext} which are trained on a huge ImageNet~\cite{imagenet} dataset as feature extractors in a common architecture between five models.
    \item Training five models containing layers pretrained on a Vision Dataset(ImageNet~\cite{imagenet}) on the IMDB-Image dataset.
\end{itemize}

The contributions of this paper are stated as follows:

\begin{itemize}
    \item Exchanging knowledge between language and vision models through transfer learning
        and data transformations.
    \item Generating an image dataset for IMDB textual dataset, avoiding domain shifts between source and target datasets (ImageNet~\cite{imagenet} and IMDB-Images) with pixel normalization.
    \item Harnessing the pre-training of vision models on a large image dataset in text classification, and promising results.
\end{itemize}

\section{Related Work}
\subsection{Transfer Learning}
 The modern learning paradigm for most Vision models, is mostly based on extracting features from the ImageNet dataset~\cite{imagenet}, then finetuning certain layers on a new smaller dataset concerned with a new task. The concept of $Transfer  Learning$ evolved from when it was first introduced by Stevo Bozinovski and Ante Fulgosi~\cite{transferlearning}~\cite{transferlearningreminder}. Due to the revolution in hardware allowing training on very large datasets, most training paradigms transfer knowledge obtained from a large dataset to solve a target task with a target dataset. Authors in \cite{compsurvey} provide a complete survey on $Transfer Learning$, its applications, types, methods, and challenges. Yosinski et al.~\cite{howtransferable} conducted a detailed study on what kind of features different layers in a Neural Network learn about the source dataset, and how $Transfer Learning$ can make the most of the acquired knowledge to solve a target task. 

\subsection{Transfer Learning in Language and Vision}
Transfer Learning in Computer Vision and Natural Language Processing as two separate sub-fields, is the current standard, as it allows large pretrained models to be used in multiple tasks after acquiring a certain understanding of features in a large dataset.

\subsubsection{Transfer Learning in Language}

BERT~\cite{Bert} representations and Fine-Tuning,
allowed for state-of-the-art results in tasks such as Text Classification, following different approaches mentioned in \cite{bertfine}, commonly using the $[CLS]$ representations from certain layers as an input to a classifier, freezing the BERT model or finetuning it depending on the task and size of task dataset. The current approaches mentioned in \cite{bertfine} achieved very low error rates. $Transfer Learning$ in this case, is mostly learning contextualized representations for the input text, in a self-supervised manner, where the generated embeddings can be directly used as an input to another model to solve a target task. 

\subsubsection{Transfer Learning in Vision}
Vision Models, more specifically \cite{alexnet}~\cite{shufflenetv2}~\cite{vgg16}~\cite{resnet}~\cite{resnext} are pretrained on ImageNet~\cite{imagenet}. Considered benchmarks for image classification, they achieved very high accuracies for smaller datasets through $Transfer Learning$ and $Domain Adaptation$~\cite{domainshift}. Being pretrained on a very large dataset (ImageNet~\cite{imagenet}), the features they can extract are useful in Machine Learning tasks on different datasets. Finetuning pretrained features on a target task often gives optimal results as discussed by Yosinski et al.~\cite{howtransferable}. 

\subsection{Text Classification using Convolutional Neural Networks}
The work conducted by Yoon Kim~\cite{kim2014convolutional} suggested a CNN based architecture with multiple filter widths and feature maps, and a fully connected layer,  using padding to produce input vectors of a fixed length. A similar approach is taken in~\cite{conv4text}, where sentence classification is achieved, using $Word2Vec$~\cite{word2vec} embeddings, 3 filter region sizes, with 2 filters for each region size, which generates 6 variable size feature maps, forming a feature vector after concatenation, then fed into a Softmax layer. The mentioned approach, treats text embeddings as fixed length inputs, with no regard to which order the tokens in the sentence appear, and no consideration of the geometrical abilities and nature of kernels in a Convolutional Neural Network.

\section{Preliminaries}
\subsection{IMDB Dataset}
IMDB~\cite{imdb} is a Polarity Dataset for Sentiment Analysis or Text Classification in broader terms, it contains 50000 Sentences and their binary class labels, being either "Positive" or "Negative", IMDB is a relatively small dataset that provides a level of flexibility and suitable testing for the study this paper is concerned with, due to the computational resources both Data Generation and Training of different models require. 
		
\subsection{BERT}
			
			BERT: Pre-training of Deep Bidirectional Transformers for Language 
					Understanding~\cite{Bert}, is a language representation model with
					the Transformer~\cite{attention} as its building block,
					pretrained on very large unlabelled textual data for two main 
					tasks: Masked Language Modelling, where the model is required to 
					predict words intentionally masked in a multi-layered context. The Second Task is Next Sentence
					Prediction, also in a self-supervised manner, BERT is trained 
					to classify a sentence as "Next" or "Not Next" for a previous 
					sentence as input. BERT pre-training takes the possible relationships between two sentences into consideration, especially when being trained on the entire Wikipedia English Corpus and the Books-Corpus~\cite{bookcorpus}. The Attention mechanism employed in all layers of BERT produces semantically rich, and 
					context-sensitive representations, where a token might have many 
					representations depending on the context it is being used in.
					For the sake of this study, A pretrained BERT model provided by
					HuggingFace~\cite{huggingface} is used, containing twelve
					layers, 768 hidden-size (Each Layer produces a 768 sized vector for 
					each word), 12 attention heads, and 110M parameters.\footnote{more information on the BERT model used is at: 	\url{https://huggingface.co/transformers/pretrained_models.html}}

\subsection{DeepInsight and t-SNE}
                Sharma et al in \cite{deepinsight}, proposed a methodology to transform
				non-image data to an image, using t-SNE~\cite{tsne} or K-PCA~\cite{kpca} to project the transpose of a dataset creating a set of features related on a 2D plane according to their similarity, re-transposing the set to obtain the original set size with [NxNx3] images as new samples. The method was heavily used in Cancer Detection and related medical applications.
                t-SNE~\cite{tsne} (t-distributed Stochastic Neighbor Embedding), is a similarity measuring and dimensionality reduction technique, developed in 2008 by Geoffrey Hinton and Laurens Van Der Maaten. What separates t-SNE~\cite{tsne} from classical dimensionality reduction techniques, is that it is a non-linear method, meaning that datasets with a very large number of dimensions can be easily viewed or projected on a 2D or 3D dimensional space, even when features are not related linearly. The mapping from the high dimensional space to the lower dimensional space (2D or 3D), happens according to the similarity between features and data points, where samples with similar features are clustered together. The similarity between two points is computed as probabilities based on Euclidean Distances between pairs of data points.

\section{Method}
In this section, the steps taken to generate the dataset used in our experimental part are explained. Going through obtaining BERT-embeddings from the pretrained BERT~\cite{Bert} model for the original IMDB dataset~\cite{imdb}, transforming the embeddings into images, and visualization of some obtained IMDB-Image Dataset samples, and normalizing the pixel space. Along with an analysis of $Source$ and $Target$ domains for the $Transfer Learning$ conducted in this paper, and the architectures trained on the generated IMDB-Image dataset, providing sample feature maps produced by the pretrained layers from \cite{resnet}~\cite{shufflenetv2}~\cite{vgg16}.

\subsection{Generating the IMDB-Image Dataset}

\subsubsection{BERT Embeddings Generation}
                A very special feature of BERT is the [CLS] token that is added at the 
				beginning of a sentence embedding at each layer output, which indicates 
				the beginning of a sentence and is also a unique Sentence Representation
				for classification purposes. Since our procedure requires a high Dimensional Space because we attempt to obtain images after a t-SNE~\cite{tsne} projection of different features representing each input from IMDB~\cite{imdb}, as it will be further discussed in following sections. The study conducted in \cite{bertvisual}, demonstrates the semantic nature of the output from the higher layers of BERT, thus, the [CLS] embeddings from the last six layers are concatenated for each input sentence from the IMDB-Dataset~\cite{imdb}, giving a [6x768] sized vector for each input sample as Fig.1 depicts.
				
				\begin{figure}[htpb]
                \centerline{\includegraphics[scale=0.7]{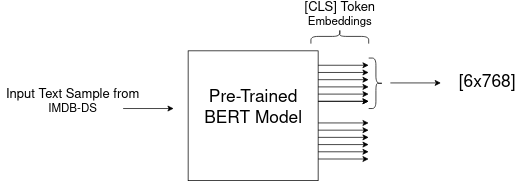}}
                \caption{IMDB [CLS] Embeddings from the Last Six layers of BERT, where the input to the pretrained model is indexes representing each word in a sentence, outputting a fixed [768] sized vector from each layer. The outputs of $[CLS]$ tokens from the last six layers are stacked into a [6X768] sized embedding for each text sample.}
                \label{figure-1}
                \end{figure}
                
\subsubsection{Transforming BERT Embeddings into Images}
                After stacking 6 vectors with 768 features for each sample in the original IMDB~\cite{imdb} dataset, our dataset is now of shape [50000, 4608]. Following the pipeline introduced in\cite{deepinsight}, We have $n=50000$ samples, with $d=4608$ features for each sample, thus our dataset can be defined as $D={\{x_1,x_2,....,x_n}\}$, where each feature vector $x$ is defined as $x=\{f_1,f_2,...,f_d\}$, our feature set is then defined as $F=\{x_1,x_2,...,x_n\}$, where each feature $f$ is a row with $n$ elements representing the number of samples containing that feature. In short terms $F=D^T$, and transposing the instance-feature matrix allows for features to be treated as elements that can be projected on a 2D plane according to similarities between samples in the dataset. The obtained 2D plane demonstrated in Fig.2, represents the location of features according to t-SNE projection of the feature-instance matrix. A Convex Hull algorithm is used to isolate the rectangle containing all the points as depicted in Fig.2. The rectangle is then rotated to obtain a horizontal matrix containing Cartesian coordinates for the pixels. 
                
                \begin{figure}[htpb]
                \centerline{\includegraphics[scale=0.55]{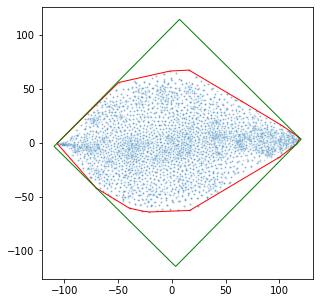}}
                \caption{Feature locations represented by blue points on a 2D plane, as described above. The green rectangle represents the smallest rectangle containing all points and is obtained using a Convex Hull algorithm. We can observe a size of [50x50] is obtained for the rectangle which is then rotated to obtain a horizontal image, ready for use in a Convolutional Neural Network.}
                \label{figure-2}
                \end{figure}
                
                Since the pixel space is limited by a fixed size, the points representing feature locations can have more than 1 feature per location due to the continuous nature of feature values. Therefore, during mapping features to their locations, averaging is required to generate a discrete space of pixels. Each feature is then mapped to its location, averaging features which fall on the same point or location on the 2D plane. Respecting our hardware capacity and to avoid excessive overlapping of features on the same location, a size of [50x50] for our pixel space is chosen as a configurable parameter in t-SNE. 
                
                \begin{figure}[htpb]
                \centerline{\includegraphics[scale=0.4]{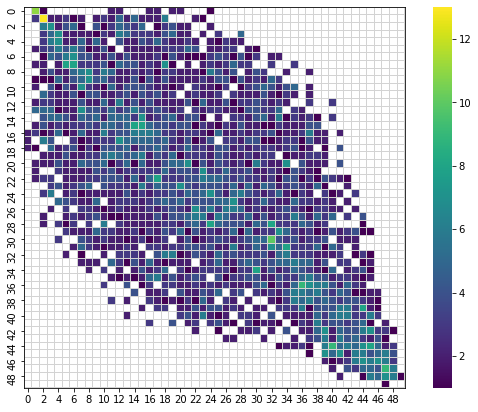}}
                \caption{A Density Matrix, after rotating the Convex containing all the points representing locations projected on a 2D plane with respect to their Cartesian Coordinates obtained from t-SNE~\cite{tsne}. Showing regions that are more dense, due to features being mapped to the same locations causing overlapping. We can observe that our new IMDB-Image Dataset has a certain geometrical distribution of features, with clear edges and blobs.}
                \label{figure-3}
                \end{figure}

              Fig.3 is an experimental result of applying t-SNE~\cite{tsne} to our dataset $D$. The Density Matrix reflects the actual distribution of features in the new IMDB-Image Dataset. \\ The next step is to map feature values to their Cartesian coordinates for each element in $D$, and averaging feature values that share the same coordinates.\newline

\subsubsection{Image Data Visualization}
                After running the method with a t-SNE~\cite{tsne} backbone on our dataset $D$, we obtain gray-scale images with height=50, width=50 and 3 
				channels having the same values for each pixel, most probably caused by the unnatural data-source which is the pretrained BERT model.
				Fig.4 demonstrates the Image-Embeddings for the first three IMDB input-samples.\\
				Text Samples for Images in Fig.4:\\
				\begin{itemize}
				    \item "One of the other reviewers has mentioned that after watching just 1 Oz episode you'll be hooked. The..."
				    \item "A wonderful little production. <br /><br />The filming technique is very unassuming- very old-time-B..." 
				    \item "I thought this was a wonderful way to spend time on a too hot summer 
				weekend, sitting in the air con..."
				\end{itemize}
				 
				\begin{figure}[htbp]
                \centerline{\includegraphics[scale=0.23]{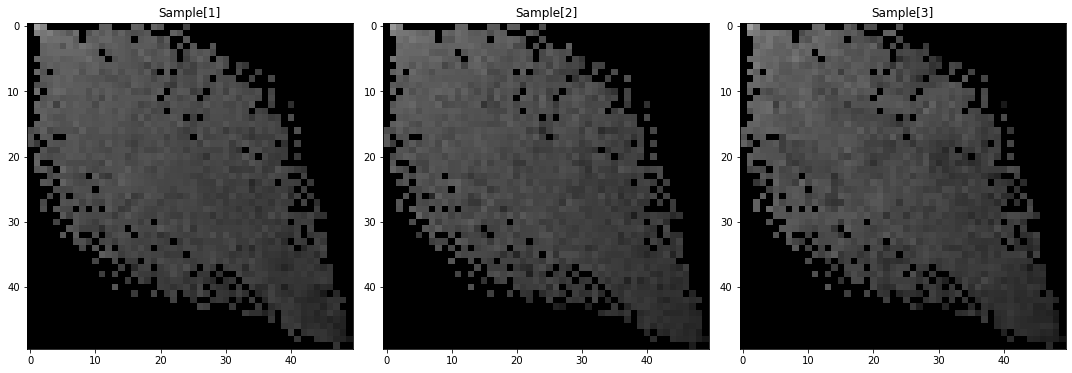}}
                \caption{Image Embeddings for the three first IMDB samples, each image is a projection of a [6x768] sized embedding obtained from BERT for each sample in the IMDB~\cite{imdb} dataset (The text samples above respectively). A new IMDB-Images dataset is generated in the same manner, giving 50000 images, each one representing a text sample in 50000 from the original IMDB~\cite{imdb} dataset.}
                \label{figure-2-1}
                \end{figure}

\subsection{Data Domains and Transfer Learning}		
				Our generated IMDB-Image dataset and the original dataset (ImageNet~\cite{imagenet}) on which
			    the models used in this work are trained, are extremely different
			    as depicted in Fig.5, which according to \cite{domainshift} causes a 
			    distribution mismatch and domain shift problems to the classifiers. 
			    Generalization across domains is extremely affected by the nature of domains and the style of the data especially in Visual Understanding related tasks.
			    
			According to \cite{compsurvey} a domain $D$
			 is composed of a feature space $\chi$ and a marginal distribution
			$P(X)$ formulated as:
			\begin{equation}
			    D = {\chi,P(X)} 
			\end{equation}

				where $X$ is an instance set which is defined as:
				$$X={x|xi \in \chi, i=1,....,n}$$

			The Task $T$ is composed of a label space $Y$ and a decision 
			function $f$, meaning:
			\begin{equation}
			    T = {(Y,f)}   
			\end{equation}
				
				where $f$ is learned explicitly via training data samples.
				
\subsubsection{Data Domains}

                \begin{figure}[htbp]
                \centerline{\includegraphics[scale=0.75]{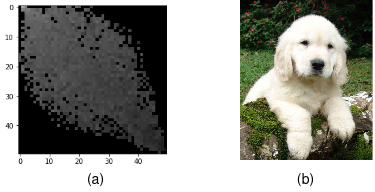}}
                \caption{(a): An image sample from the generated IMDB-Images dataset, showing a gray scale distribution with weak edges and no obvious geometrical features or shapes, (b): An image of a Dog from the $Source$ Dataset (ImageNet~\cite{imagenet}), an RGB image with clear geometrical features and shapes. This figure depicts the differences between the $Source$ and $Target$ datasets, and suggests inevitable domain shifts between the two.}
                \label{figure-3-1}
                \end{figure}
                
                Fig.5 shows obvious differences in data style in the source and target datasets, which according to \cite{lookatdata} and \cite{improving} and further discussed in \cite{decadesurv} causes a distribution mismatch and domain shifts, due to the differences between the two domains like channels, colors, background,
				lighting, etc. This is a major problem in Transfer Learning between 
				image datasets. Survey \cite{domainshift} gives a broad overview of the recent approaches and methods developed for Transfer learning to overcome the mentioned issues through Domain Adaptation.

\subsubsection{Transfer Learning}
                For successful Transfer Learning to be achieved, an architecture should
				be able to adapt the Target Domain $D_T$ to Source Domain
				$D_S$, which are IMDB-images and ImageNet~\cite{imagenet} respectively in our case. One special technique for domain adaptation so far proposes specific  training for domain prediction \cite{ganin2016domain}.
				Limited by compute power and extreme domain shifts, in this work, another
				approach is taken.
				In \cite{howtransferable}, Jason Yosinski et al, stress on the different kinds of features different zones of layers learn in a neural network, stressing on the generality observed in lower layers and specificity in higher layers, meaning general features like curves, color blobs, and edges are extracted in the lower layers, and task specific features are handled by higher layers. 
				Threatened by the large model sizes in our pretrained arsenal~\cite{alexnet}~\cite{shufflenetv2}~\cite{vgg16}~\cite{resnet}~\cite{resnext}, and the relatively small dataset with only 50000 samples, which could lead to extreme overfitting, the approach taken in this work is to focus on features that both datasets share instead of forcing the 
				model to learn the domains themselves as targets. Since the color
				channels are clearly going to cause a major domain shift, the focus should
				be on geometrical features like edges, curves and blobs. In order to 
				have more defined edges, normalization of the entire pixel space is 
				applied, a normalization technique named $Z- Normalization$~\cite{normimproving}, adjusting image contrast after moving our input images to a clearer pixel space:
				\begin{equation}
				\begin{split}
				        &\mu = \mathbb{E}_{x} \in\mathcal{X}[x],\\
				        &\sigma^2 = \mathbb{E}_{x}\in\mathcal{X} [(x-\mu)^2],\\
				        &\hat{x}_i = \frac{x_i - \mu}{\sigma + \epsilon}
			    \end{split}
				\end{equation}
				       
				$\mathcal{X} = [x_{1}, x_{2}, ... , x_{n}]$ is a set of input vectors,
				$\mu$, $\sigma$ are the mean and standard deviation of the entire image pixel space, 
				$\epsilon$ is a small value to prevent dividing by zero or small denominators.
				
				\begin{figure}[htbp]
                \centerline{\includegraphics[scale=0.7]{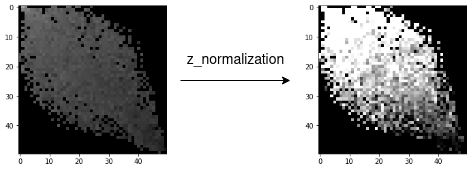}}
                \caption{The image to the left, represents a raw image from the generated IMDB-Images dataset. The image to the right, is the same as the one to its left after $Z-Normalization$, showing stronger edges and more defined geometrical shapes like blobs.}
                \label{figure-4}
                \end{figure}
                
                As Fig.6 shows, the mentioned normalization technique prevails in creating more distinguishable feature zones
                in one image, due to the new mean which is close to 0.0 and standard deviation nearing 1.0, which imposes a standard normal distribution on the features, removing noisy regions and enhancing outlying pixels.

\subsection{Architectures used}

                Since our IMDB-Image dataset is very small compared to the source 
				ImageNet dataset~\cite{imagenet}, the Convolutional Feature Extractors are sliced from their original pretrained models, and stacked to a Convolutional Auto-Encoder with randomly initialized parameters, followed by a Dense (Linear) Classifier. A common approach would be to freeze the pretrained feature extractors, which was followed in this paper to avoid any overfitting problems. We also note that the Convolutional Auto-Encoder was designed to keep the same activation flow based on the pretrained feature extractors, and thus keeps a similar choice of activation functions and initialization.
                
                \begin{figure}[htbp]
                \centerline{\includegraphics[scale=0.70]{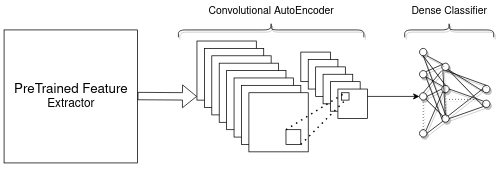}}
                \caption{Main Architecture Used, the pretrained block represents early layers from five pretrained Vision Models~\cite{alexnet}~\cite{shufflenetv2}~\cite{vgg16}~\cite{resnet}~\cite{resnext}, outputting the input to a Convolutional Auto-encoder, stacked to a Dense Classifier (3 Linear fully connected layers)}
                \label{figure-5}
                \end{figure}
                
				As it is depicted in Fig.7, the most important part of the 
				architecture is the pretrained feature extractor. In this paper, for 
				the sake of comparison and confirmation, early layers from five 
				pretrained models were used as feature extractors, followed by the 
				exact same Conv-AE (Convolutional AutoEncoder) and Dense classifier to ensure fairness in comparison.
				The detailed architectures of the pretrained vision models are outside the scope of this paper.
		
\subsubsection{pretrained Feature Extractors}\footnote{All pretrained models can be found at: \url{https://pytorch.org/vision/stable/models.html}}

The following list represents the number of layers sliced from the pretrained vision models. We note this choice has been immensely affected by the available hardware, and also by repetitive visual assessment of the feature-maps outputted by several combination of layers for each one.
\begin{itemize}
			 \item{AlexNet: introduced in \cite{alexnet}, Using the first two pretrained Convolutional Layers, outputs
						192 feature maps for each input image from the IMDB-image dataset, 
						with fairly distinguishable differences and focus.}

			 \item  ResNet: A deep residual model from \cite{resnet}, known as wide-resnet50-2. We used the first downsampling Convolutional layer and the first residual layer.

			 \item  ResNext: \cite{resnext},
						proposes an aggregated version of the previous ResNet. For
						the feature extractor we need in this experiment, the first Convolutional layer,
						and the first Residual layer are used as well.

			 \item ShuffleNet V2: From \cite{shufflenetv2} 
							the first Convolutional layer followed by Batch normalization,
							and stage2 mentioned in the paper. 

			 \item VGG16: introduced in \cite{vgg16},
					We only use the first 12 layers, containing 4 Convolutional layers for the feature extractor in this experiment.
			 
\end{itemize}
Fig.8 shows sample Feature maps from the feature extractors from:
				resNet~\cite{resnet},. ShuffleNet~\cite{shufflenetv2}, Vgg16~\cite{vgg16}.
				
			    \begin{figure}[hbt!]
                \centerline{\includegraphics[scale=0.8]{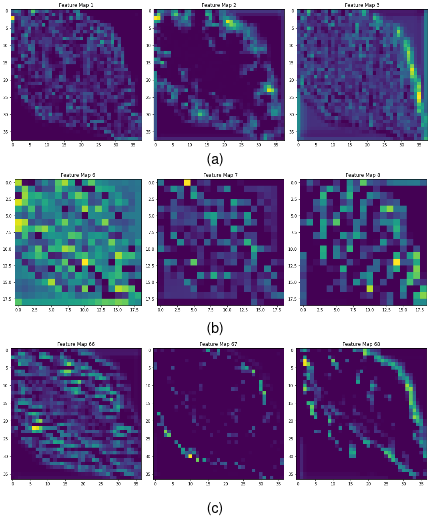}}
                \caption{Feature Maps from early layers of (a) : resNet, (b) : shuffleNet, (c) : VGG16. The pretrained feature extractors used in our common architecture do in fact output feature maps with defined geometrical features extracted due to the pre-training on the ImageNet~\cite{imagenet} dataset. We can observe the adjusted edges and blobs using $Z-Normalization$ being successfully detected by all pretrained feature extractors, clearest in the depicted three.}
                \label{figure-6}
                \end{figure}

In Fig.8 we can observe that the pretrained models are in fact able to extract global features from the new dataset, edges and curves are distinguished by all the pretrained feature extractors, but clearest in the depicted three. Yielding that \cite{howtransferable} stated a concrete study of the transfer-ability of 
pretrained models to other datasets and tasks, requiring a considerate treatment of the nature of the features learned by different layers, and the different natures and domains of the source and target datasets.

\section{Experiment}
This section discusses the training procedure for the mentioned architecture containing five different pretrained feature extractors from the Vision Models~\cite{alexnet}~\cite{shufflenetv2}~\cite{vgg16}~\cite{resnet}~\cite{resnext}, pretrained on ImageNet~\cite{imagenet}, and the experimental results achieved by each one on the generated IMDB-Image dataset.

\subsection{Training}
            As shown in Fig.7, our five models share a Convolutional AutoEncoder, and three Linear Layers, and differ in the pretrained feature extractors obtained from 5 different pretrained vision models~\cite{alexnet}~\cite{shufflenetv2}~\cite{vgg16}~\cite{resnet}~\cite{resnext}. Given the differences in data domains depicted in Fig.5, avoiding any possible dataset and covariate shifts is necessary. Using $ReLU$ to keep the same activation patterns within layers of our architecture, given that all 5 pretrained feature extractors contain $ReLU$ activations. Batch Normalization~\cite{batchnorm} is applied after each convolutional layer in our Convolutional AutoEncoder block, since our pretrained feature extractors are trained with 1000 classes on the label side, and our classification task only has 2 classes, this causes an Internal Covariate Shift, representing the change in the distributions of internal nodes of our network as mentioned in \cite{batchnorm}. Adam Optimizer~\cite{Adam}, is used for all 5 models, to ensure fairness in comparisons, and a fixed batch size of 32. Our IMDB-Image dataset contains 50000 images samples,
            we split the dataset into a 40000 train and a 10000 validation samples. Differential learning rates are used for our Adam~\cite{Adam} optimizer, while keeping the pretrained feature extractors frozen during training, since our dataset is very small compared to the $Source$ Dataset (imageNet~\cite{imagenet}).\\
			An NVIDIA GEFORCE GTX 1060 GPU was used for the entire procedure. 
\subsection{Experimental Results}
			As mentioned in the previous subsection, the different models were trained,
			with different learning rates, but yet reached very close Validation Accuracies. Higher learning rates caused all models to stagnate and converge very fast to local optimal points, which prevented the models to further learn features necessary to distinguish between different variations contained in each image representing the text samples in the original IMDB-dataset~\cite{imdb}.
			
			The following table depicts the number of feature maps outputted by each Feature Extractor along with different Learning Rates used for each one, and the corresponding achieved Validation Accuracies. Fig.9 shows the progress of the validation performance of the five models during training.
            TABLE 1 and Fig.9 both emphasize the very close results obtained from training all five models on the same IMDB-Image dataset.
        
            \begin{table}[htbp]
            \centering
            \caption{Nbr of FM's (Number of feature-Maps), Learning Rates for CONV-AE (Convolutional AutoEncoder and LC (Linear Classifier) and Val Acc (Validation Accuracy) for each model defined by its Feature Ext (Extractor)}
            \begin{tabular}{c c c c c c c c c}
            \hline\\
            Feature Ext &   Nbr of FM's & CAE LR  & LC LR &   Val Acc  \\\\ 
            \hline\\
            AlexNet~\cite{alexnet}   & 192 &0.00001 & 0.0005 &  0.87 ($\pm$0.01)  \\\\
            ResNet~\cite{resnet}    & 256 &0.00005 & 0.0001 &  0.85 ($\pm$0.01)  \\\\
            ResNext\cite{resnext}     & 256 &0.00005 & 0.001  &  0.85 ($\pm$0.01)  \\\\
            ShuffleNet~\cite{shufflenetv2}  & 116 &0.0005  & 0.001  &  0.86 ($\pm$0.01)  \\\\
            VGG16~\cite{vgg16}     & 256 &0.00005 & 0.001  &  0.86 ($\pm$0.01) \\\\
            \hline\\
            \end{tabular}
            \end{table}	
\begin{figure}[ht]
    \centering

\begin{tikzpicture}[scale = 0.8]
\label{second-meth curves}

\begin{axis}[
    title={},
    xlabel={Epoch},
    ylabel={Validation Accuracy (\%)},
    xmin=1, xmax=15,
    ymin=0, ymax=100,
    xtick={1,2,3,4,5,6,7,8,9,10,11,12,13,14,15,16},
    ytick={0,20,40,60,80,100},
    legend pos=south east,
    ymajorgrids=true,
    grid style=dashed,
]

\addplot[
    color=blue,
    mark=square,
    ]
    coordinates {
    (1,10)(2,30)(3,51)(4,60)(5,70)(6.72)(7,76)(8,78)(9,81)(10,83)(11,84)(12,85)(13,86)(14,86.5)(15,87)
    };
    \addlegendentry{AlexNet-based}
    
\addplot[
    color=red,
    mark=square,
    ]
    coordinates {
    (1,5)(2,24)(3,49)(4,53)(5,60)(6.64)(7,70)(8,73)(9,79)(10,80)(11,82)(12,83)(13,84)(14,84.5)(15,85)
    };
    \addlegendentry{ResNet-based}

\addplot[
    color=green,
    mark=square,
    ]
    coordinates {
    (1,7)(2,22)(3,44)(4,51)(5,60)(6.61)(7,69)(8,72)(9,77)(10,79)(11,81)(12,81)(13,83)(14,84.5)(15,85)
    };
    \addlegendentry{ResNext-based}

\addplot[
    color=purple,
    mark=square,
    ]
    coordinates {
    (1,8)(2,20)(3,41)(4,59)(5,60)(6.61)(7,66)(8,70)(9,75)(10,79)(11,82)(12,82.5)(13,83.2)(14,85)(15,86)
    };
    \addlegendentry{ShuffleNet-based}

\addplot[
    color=black,
    mark=square,
    ]
    coordinates {
    (1,11)(2,21)(3,40)(4,56)(5,59)(6.62)(7,65)(8,71)(9,74)(10,76)(11,81)(12,82)(13,83.5)(14,84)(15,86)
    };
    \addlegendentry{ShuffleNet-based}

\end{axis}
\end{tikzpicture}
\caption{Validation Accuracies VS. Epochs for the five models}
\end{figure}
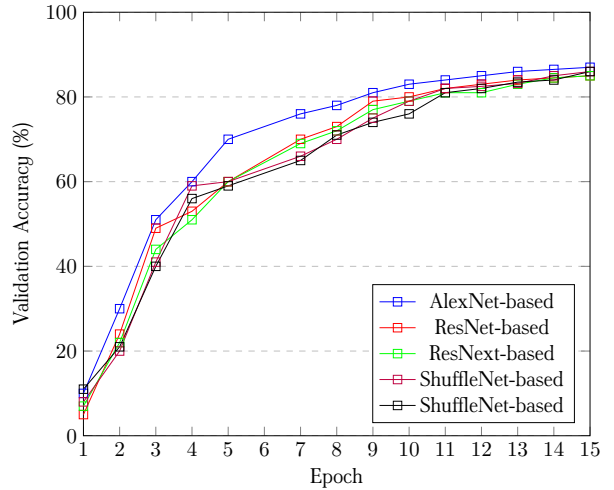

Freezing the pretrained Feature Extractors allows for the complexity of the model as a whole for the five variations to drop, avoiding any chance of overfitting due to the gray scale nature of our IMDB-Image dataset and the RGB channel space of the $Source$ Dataset (ImageNet~\cite{imagenet}).
The experimental analysis conducted in this paper, is explicitly dedicated to the results obtained by our method and architectures.\newline
Given that the pretrained Feature Extractors, followed by the exact same Convolutional AutoEncoder and Dense Classifier, are trained on the same generated IMDB-Image dataset, and observing the experimental results obtained in TABLE 1 and Fig.9, suggests the following: 

\begin{itemize}
    \item The normalized IMDB-Image Dataset, is still not fully avoiding domain shifts due to its raw gray-scale nature.
    \item The different number of features outputted by each feature extractor, and the almost identical Validation Accuracies, suggests that some feature maps are duplicated, also due to the gray scale nature of the IMDB-Image dataset.
    \item We can clearly see in Fig.9 that the five models learned at different rates, yet reached close Validation results, this can imply that the models do indeed vary in complexity and generalization abilities, yet limited by dataset size.
    \item The general features learned from the early layers used as feature extractors are extremely similar as suggested in \cite{howtransferable}, yielding similar results even with different learning rates, since the feature maps play the role of fixed representations for the same dataset.
    \item Although the Validation Results accomplished do not rise to the State-Of-The-Art in Text Classification, they are still promising given the limited data size used..

\end{itemize}

\section{Data and Code Availability}
The code for:
\begin{itemize}
    \item Used IMDB dataset can be found at \href{https://www.kaggle.com/lakshmi25npathi/imdb-dataset-of-50k-movie-reviews}{IMDB}.
    \item Code for generating IMDB-Image Dataset.
    \item BERT Embedding, Data Transformation and loading.
    \item Different architectures and training scripts using PyTorch.
    \item Reproducible paradigm with commented and explained steps.
\end{itemize}

Are available and can be accessed at \url{https://github.com/EddCBen}.\newline
\newpage
\section{Conclusion}

  In this paper a new approach to Text Classification via Supervised Learning is suggested, using Transfer Learning of pretrained Vision Models on an Image Dataset (ImageNet~\cite{imagenet}), to a textual polarity Dataset (IMDB~\cite{imdb}) embedded using a pretrained BERT~\cite{Bert} model, where text embeddings are transformed into images using t-SNE~\cite{tsne} feature similarity measuring (Inspired from the work done in DeepInsight~\cite{deepinsight}), and pretrained Vision models, are used as feature extractors for a much smaller Neural Classifier to learn sentiment labels.
  The contributions of our work, are mainly the generation of a new dataset representing textual data from the IMDB~\cite{imdb}, avoiding possible domain shifts with pixel normalization, and achieving text classification using pretrained layers for vision tasks. Future work aims to further normalize the discussed approach, as it gives promising results for a small dataset, opening a new challenge for the fusion of Language and Vision via Transfer Learning and Data transformation.

\bibliography{refs.bib}
\bibliographystyle{plain}
\end{document}